\documentclass[journal]{IEEEtran}
\usepackage{graphicx}
\usepackage{color}
\usepackage[numbers,sort&compress]{natbib}
\usepackage{algorithm,algpseudocode,algorithmicx}
\usepackage{amsmath,latexsym,amssymb,amsthm,array,amsfonts,algorithm,algpseudocode,booktabs,graphicx,subfigure,multirow,cuted,stfloats}
\ifCLASSINFOpdf
\else
\fi
\begin{document}
\title{Adaptive Prototypical Networks with Label Words and Joint Representation Learning for Few-Shot Relation Classification}

\author{Yan~Xiao,
        Yaochu~Jin,~\IEEEmembership{Fellow,~IEEE,}
        and~Kuangrong Hao% <-this % stops a space
\thanks{Y. Xiao and K. Hao are with the Engineering Research Center of Digitized Textile \& Apparel Technology, Ministry of Education, College of Information Science and Technology, Donghua University, Shanghai 201620, China. Email: xiaoyan@mail.dhu.edu.cn; krhao@dhu.edu.cn}% <-this % stops a space
\thanks{Y. Jin is with the Engineering Research Center of Digitized Textile \& Apparel Technology, Ministry of Education, College of Information Science and Technology, Donghua University, Shanghai 201620, China. He is also with the Department of Computer Science, University of Surrey, Guildford, Surrey GU2 7XH, United Kingdom. Email: yaochu.jin@surrey.ac.uk. (\textit{Corresponding author})}
\thanks{Manuscript received xxxx, 2021; revised xxxx, 2021.}}


\markboth{Journal of \LaTeX\ Class Files,~Vol.~xx, No.~x, August~20xx}%
{Shell \MakeLowercase{\textit{et al.}}: Bare Demo of IEEEtran.cls for IEEE Journals}



\maketitle

\begin{abstract}

Relation classification (RC) task is one of fundamental tasks of information extraction, aiming to detect the relation information between entity pairs in unstructured natural language text and generate structured data in the form of entity-relation triple. Although distant supervision methods can effectively alleviate the problem of lack of training data in supervised learning, they also introduce noise into the data, and still cannot fundamentally solve the long-tail distribution problem of the training instances. In order to enable the neural network to learn new knowledge through few instances like humans, this work focuses on few-shot relation classification (FSRC), where a classifier should generalize to new classes that have not been seen in the training set, given only a number of samples for each class. To make full use of the existing information and get a better feature representation for each instance, we propose to encode each class prototype in an adaptive way from two aspects. First, based on the prototypical networks, we propose an adaptive mixture mechanism to add label words to the representation of the class prototype, which, to the best of our knowledge, is the first attempt to integrate the label information into features of the support samples of each class so as to get more interactive class prototypes. Second, to more reasonably measure the distances between samples of each category, we introduce a loss function for joint representation learning to encode each support instance in an adaptive manner. Extensive experiments have been conducted on FewRel under different few-shot (FS) settings, and the results show that the proposed adaptive prototypical networks with label words and joint representation learning has not only achieved significant improvements in accuracy, but also increased the generalization ability of few-shot RC models.

\end{abstract}

\begin{IEEEkeywords}
Relation classification (RC), few-shot learning (FSL), adaptive prototypical networks.
\end{IEEEkeywords}

% \fi
\IEEEpeerreviewmaketitle

\section{Introduction}

% \usepackage{multirow}
\begin{table*}[]
 \centering
 \caption{A data example of a 3-way 2-shot scenario of FSRC. The head entity and tail entity are indicated by red and blue respectively for each sentence. The original label words and descriptions of each category are also given in the left column of the table. The correct relation class for the query instance is R2.}
\begin{tabular}{ll}
\hline
\multicolumn{2}{c}{\textbf{Support Set}}                                                                                                                                                                                                                                                                                                                          \\ \hline
\multicolumn{1}{l|}{\multirow{2}{*}{\begin{tabular}[c]{@{}l@{}}R1: country of citizenship (the object is a country \\ that recognizes the subject as its citizen)\end{tabular}}}          & \begin{tabular}[c]{@{}l@{}}Instance1: \textcolor[rgb]{0.00,0.50,1.00}{Chris Huffins} (born 15 April 1970) is an athlete from the \textcolor[rgb]{1.00,0.00,0.00}{United States} who competed \\ in the field of Decathlon.\end{tabular} \\
\multicolumn{1}{l|}{}                                                                                                                                                                     & Instance2: During the Three Kingdoms era, \textcolor[rgb]{0.00,0.50,1.00}{Liu Bei}'s \textcolor[rgb]{1.00,0.00,0.00}{Shu} was based in Sichuan.                                                                                         \\ \hline
\multicolumn{1}{l|}{\multirow{2}{*}{\begin{tabular}[c]{@{}l@{}}R2: religion (religion of a person, organization or \\ religious building, or associated with this subject)\end{tabular}}} & Instance1: He founded a literary prize which in 1901 was won by \textcolor[rgb]{1.00,0.00,0.00}{Ndoc Nikaj}, a \textcolor[rgb]{0.00,0.50,1.00}{Catholic} priest.                                                                        \\
\multicolumn{1}{l|}{}                                                                                                                                                                     & \begin{tabular}[c]{@{}l@{}}Instance2: These 67 used an old candle from prominent \textcolor[rgb]{0.00,0.50,1.00}{Unitarian} preacher \textcolor[rgb]{1.00,0.00,0.00}{William Ellery Channing}\\ to supply the flame.\end{tabular}       \\ \hline
\multicolumn{1}{l|}{\multirow{2}{*}{\begin{tabular}[c]{@{}l@{}}R3: developer (organisation or person that developed \\ the item)\end{tabular}}}                                           & Instance1: \textcolor[rgb]{0.00,0.50,1.00}{MicroIllusions} published \textcolor[rgb]{1.00,0.00,0.00}{The Faery Tale Adventure} first in 1986.                                                                                           \\
\multicolumn{1}{l|}{}                                                                                                                                                                     & \begin{tabular}[c]{@{}l@{}}Instance2: Since \textcolor[rgb]{1.00,0.00,0.00}{tcsh} was based on the csh code originally written by \textcolor[rgb]{0.00,0.50,1.00}{Bill Joy}, it is not considered \\ a clone.\end{tabular}              \\ \hline
\multicolumn{2}{c}{\textbf{Query Instance}}                                                                                                                                                                                                                                                                                                                       \\ \hline
\multicolumn{1}{l|}{R1, R2 or R3}                                                                                                                                                         & In 1689, Konstanty was one of the judges who sentenced \textcolor[rgb]{1.00,0.00,0.00}{Kazimierz} to death for \textcolor[rgb]{0.00,0.50,1.00}{atheism}.                                                                                \\ \hline
\end{tabular}
\label{tab:tab1}
\end{table*}

\IEEEPARstart{A}{S} the core task of text information extraction, relation classification (RC) is essential to text understanding, which has found many applications in natural language processing (NLP), such as text mining, question answering and knowledge graphs. The main purpose of RC is to generate relational triples $<e_{1},r,e_{2}>$ from plain natural language text, where $e_{1}$ and $e_{2}$ are entities and $r$ implies the relation between the two entities in a sentence. For example, the instance “\emph{Beijing is the capital of China.}” expresses the relation “\emph{capital of}” between the two entities “\emph{Beijing}” and “\emph{China}”.

With the continuous development of deep learning technologies, conventional methods of RC have been gradually replaced by neural network methods, among which the most popular and effective paradigm is the supervised RC. Since convolutional neural networks (CNN) based methods \cite{zeng2014relation} that automatically capture relevant lexical and sentence level features were first proposed for supervised RC task, subsequent research based on neural networks has gradually prevailed and achieved the state-of-the-art. However, as neural networks are data-hungry methods, the performance of supervised RC methods heavily depends on the quality and quantity of the training data, meaning that these methods suffer from the lack of large-scale training data sets.

In order to address the above issue, Mintz et al. \cite{mintz2009distant} proposed distant supervision relation extraction (DSRE) that can automatically generate training data via aligning triples in knowledge graphs (KGs) with sentences in corpus. To be specific, the method assumes that if a pair of entities has a relation in KGs, all sentences that contain these two entities will describe the same relation. In this way, DSRE effectively alleviates the problem of manually labeling large-scale data set, but it inevitably introduces noise into the dataset because of the strong hypothesis it makes. Therefore, most current research has been devoted to solving this problem through various methods such as multi-instance learning and attention mechanism. However, there are still a large number of categories with very few samples in DSRE, resulting in a clear long-tailed distribution in the entire dataset. For those categories with few samples, the classification performance will be dramatically deteriorated, which is a serious problem that most existing DSRE methods neglect. In fact, in real life, there may exist a situation where only few or even zero samples are available due to the complexity of operations, in which we humans indeed have the ability to learn knowledge from a few examples or even a single sample \cite{carey1978acquiring}. In other words, humans are able to learn quickly and make inferences by analogy, which inspired researchers to propose the concept of few-shot learning (FSL). The main challenge of FSL is how to produce models that can generalize from a small amount labeled data to understand new categories. A new category here means that the relation category to which the test sample belongs and has never appeared during the training. Through the training mechanism of FSL, the model combines the generalized knowledge learned from the basic data with a limited number of training samples of the new category, achieving the purpose of rapid knowledge learning.

In the past few years, research on FSL originated mainly from the field of computer vision (CV), and the commonly used data sets, Omniglot and Mini-ImageNet, are both image classification data sets. One most commonly used method for solving few-shot tasks is transfer learning \cite{pan2009survey}, which aims to learn new classes more quickly by making use of knowledge learned from basic classes with sufficient samples. Transfer learning is mainly realized by fine-tuning the weights of the pretrained network. Based on a similar principle, the recently proposed meta-learning method \cite{finn2017model} works by learning a parameterized function, which embeds various learning tasks and then can be generalized to new tasks. Another simple and effective method is the metric learning method, where the core idea is to model the distance distribution among all samples in the entire data set and then classify them according to different metrics. Among them, the most representative approach is the prototypical networks, which is based on the assumption that there exists a prototype for each class, and has achieved the state-of-the-art performance on most FS benchmarks. Even though FSL has developed very rapidly in recent years, most methods introduced above are derived from the field of CV. So far, FSL is still a young research field, especially when it is applied to the field of NLP. Compared with FSL in other fields, the RC task has its own unique challenges due to the fact that text is more diverse and complex than images, making it very difficult to directly generalize existing FSL models to NLP applications. Consequently, it is of great significance to extend and adapt FSL techniques to NLP tasks such as RC. To promote the research of FSL in the field of NLP, Han et al. \cite{han2018fewrel} first introduced FSL to the task of relation extraction and constructed a few-shot relation classification (FSRC) dataset FewRel, aiming to predict the relation for a pair of entities in a sentence by training with a few labeled examples in each relation. Table \ref{tab:tab1} shows a part of the dataset, which is an episode of data in 3-way 2-shot scenarios. In the left column, two parts of label information are shown respectively, namely the specific label words and label description (enclosed in parentheses) of each category. The purpose of the FSRC task is to determine which category the query instance of the last line belongs to.

In this work, we propose adaptive prototypical networks with label words and joint representation learning based on metric learning for FSRC, which performs classification by calculating the distances in the learned metric space. Considering that only few samples are available in the FSRC, our motivation is to maximize the advantage of existing data information. On the one hand, we augment the metric-based prototypical networks with an adaptive mixture mechanism to get a more precise class prototype, where a convex combination is formed by incorporating the label information. To be specific, not only the samples from each category, but also the label words of each category are taken into account in an adaptive convex combination of two homologous embedding spaces. On the other hand, a joint training algorithm consisting of a classification loss function and a representation learning loss function is proposed to achieve the effective representation of each sample. The purpose of the representation learning loss is to make the distance between samples of the same category much greater than that of different types in a space, which is inspired by the margin triplet loss in the face recognition task \cite{schroff2015facenet}. Consequently, not only the problem of feature sparsity is effectively alleviated, but also the performance degradation of models with fewer samples can be substantially improved, thereby improving the classification performance. Moreover, in order to fully verify the effectiveness of our proposed model, we conducted several experiments on the FewRel dataset, which is derived from real-world corpus Wikipedia and Wikidata. Experimental results demonstrate that our model achieves significant performance over the state-of-the-art baseline methods.

To sum up, the main contributions of this work are as follows:
\begin{itemize}
\item We propose an adaptive label information mixture mechanism based on the prototypical networks. To the best of our knowledge, this is the first time that information of each class has been integrated into the class prototypes for FSRC, which effectually alleviates the problem of model performance degradation due to data paucity.
\item A new joint learning method is designed for FSRC tasks, which allows the model to perform effective classification training assisted by good representation ability, thereby improving the generalization ability of the model on recognizing long-tail relations.
\item Experiments are conducted to compare the proposed algorithm with its ablation variants as well as with the state-of-the-art. Our comparative results show that the two mechanisms we introduced both play an important role for the proposed model to deliver significant performance gains over the state-of-the-art baseline methods. 	

\end{itemize}

The remainder of the paper is structured as follows. In Section II, we review the related work on relation extraction, few-shot learning and the triplet loss adopted in this work. Section III presents the specific definition of the FSRC task. In Section IV, we give the details of our proposed model. The dataset and implementation details are described in Section V, and we further analyze our model and compare it with state-of-the-art models in Section VI. Conclusions and future work are summarized in Section VII.

\section{Related work}

In this section, we briefly review the relevant background of relation classification, few-shot learning and the triplet loss function.

\subsection{Relation classification}
Relation classification (RC) is designed to select a proper relation from a predefined relation set when two target entities in the given text have been marked. Before neural networks were applied in the field of NLP, some conventional methods had achieved pretty good results in RC, such as kernel based methods \cite{bunescu2005shortest,zelenko2003kernel} and feature based methods \cite{kambhatla2004combining}. With the development of deep learning, several neural network-based methods, including convolutional neural networks (CNNs) \cite{qin2016empirical,wang2016relation} and the long short term memory (LSTM) \cite{geng2020semantic,zhou2016attention}, and Transformer \cite{soares2019matching}, have constantly refreshed the record on supervised RC tasks. Since the supervised RC task often suffers from a lack of annotated data sets, the idea of distant supervision relation extraction (DSRE) \cite{mintz2009distant} was proposed to alleviate this problem, which mainly uses the existing database to automatically align the text. In order to solve the noise problem caused by the strong assumption in DSRE, a piece-wise CNN (PCNN) model based on the position of two entities was proposed in combination with an attention mechanism \cite{lin2016neural}. Besides, a Transformer block based model was explored to capture the structure and contextual information of the whole text to distinguish the useful and the noisy information in sentences \cite{xiao2020hybrid}. Both reinforcement learning techniques \cite{feng2018reinforcement} and the generative adversarial training methods \cite{qin2018dsgan} were utilized to select high-quality sentences to reduce the effect of noise. Although there has been plenty of research dedicated to the RC task, the grand challenge of long-tail distribution datasets in RC remains to be solved, regardless of supervised RC or DSRE.

\subsection{Few-shot relation classification}

Few-shot learning (FSL) \cite{wang2020generalizing} has recently attracted increasing attention, as it enables to generalize to new classes in classification tasks by training a classifier with only a few instances. Knowledge transfer between associative categories is a key method for FSL \cite{caruana1995learning,bengio2012deep} in early years, which is achieved mainly through pre-training methods to transfer the latent knowledge of the previously learned category to the target domain. Recently, meta-learning was proposed to realize the idea of fast learning, where the simple neural attentive learner (SNAIL) meta-learning model \cite{mishra2017simple} uses the time convolutional neural network and attention mechanism to quickly learn from previous experience. Additionally, Munkhdalai and Yu \cite{munkhdalai2017meta} proposed a meta-network (MetaNet) to learn meta-knowledge across tasks and transformed its inductive bias through rapid parameterization to achieve rapid generalization. Meanwhile, the metric based FSL has attracted a lot of interest due to its simplicity and effectiveness, the basic idea of which is to learn a metric space that maps similar samples close and dissimilar ones distant so that classification can be easily performed. Based on this idea, the Siamese neural network was proposed in \cite{koch2015siamese} that employed an unique structure to learn a similarity metric from the data, while Vinyals et al. \cite{vinyals2016matching} explored matching networks with the idea of attention and external memories to map the data to obviate the need for fine-tuning when adapted to new class types. Snell et al. \cite{snell2017prototypical} defined the prototypical networks that usually classify the sample and its nearest class prototype into one category, where the class prototype is obtained by averaging all samples in each category. Moreover, a generalization of the above three models is proposed in \cite{garcia2017few}, where a graph neural network architecture is utilized to process general information in FSL tasks.

Although most existing FS methods were developed in the CV field \cite{bai2020class,lu2020robust,lai2020learning}, their success has inspired researchers to explore the application of FSL to NLP. With respect to the few-shot relation classification, Han et al. \cite{han2018fewrel} organized the FewRel dataset and adapted part of the methods above on FSRC, where the prototypical networks are pretty easy to implement and understand. Afterwards, hybrid attention-based prototypical networks for noisy FSRC is proposed in \cite{gao2019hybrid} and hierarchical attention prototypical networks for FS text classification is proposed in \cite{sun2019hierarchical}. However, one issue that has troubled most approaches is that the performance of FSL will seriously degrade when the amount of samples is small. To alleviate this problem, this work proposes an adaptive label information mixture mechanism with the joint representation learning inspired by the triplet loss to improve the generalization of FS model.

\subsection{Triplet loss in face recognition}
\label{sec:Triplet Loss in Face Recognition}
Representation learning \cite{bengio2013representation}, also known as feature learning, generally refers to the model that automatically learns the input data to obtain the features that are more representative. The large margin principle has played a key role in the course of feature learning history, producing remarkable theoretical and empirical results for classification and regression problems \cite{elsayed2018large}. Its desirable benefits include its better generalization properties and robustness to input perturbations. A large body of research \cite{sokolic2017robust,liu2016large,wang2018large} has explored the benefits of encouraging a large margin in the context of deep networks, which was realized by defining the following triplet loss function \cite{schroff2015facenet}:

\begin{align}\label{1}
L=\sum_i^N{\left[ \lVert f\left( x_{i}^{a} \right) -f\left( x_{i}^{p} \right) \rVert _{2}^{2}-\lVert f\left( x_{i}^{a} \right) -f\left( x_{i}^{n} \right) \rVert _{2}^{2}+\eta \right] _+}
\end{align}
According to Eq. (\ref{1}), given an image $x_{a}^{i}$ of a specific person as anchor, the triplet loss optimizes the embedding space by adding a margin $\eta$ in the objective function such that data points $x_{p}^{i}$ with the same identity (positive) are closer to each other than those with different identities $x_{n}^{i}$ (negative), which allows us to perform end-to-end representation  learning between the input image and the desired embedding space. However, generating all possible triplets will not contribute to the training, but result in a slower convergence instead. Therefore, an essential part of the triplet loss is the mining of triplets, and a variant of the triplet loss \cite{hermans2017defense} was utilized to implement deep metric learning in person re-identification task afterwards. It outperformed most other methods by a large margin for models trained from scratch as well as pretrained ones. Similarly, we also have made series of modifications to triplet loss, and finally got the effective combination of its application in FSRC, which we will introduce in detail in Section \ref{sec:Adaptive joint training}.

\section{Task Definition}
Suppose that $D$ is the entire labeled data, which is usually divided into two parts according to the relation category: $D_{train}$ and $D_{test}$. In other words, the two datasets have different label spaces and they are disjoint with each other. Generally, most of the current FSL algorithms employ the “episode” training strategy \cite{vinyals2016matching}, which has been proved to be practical. To this end, $D_{test}$ is further split into two parts: $D_{test-support}$ and $D_{test-query}$. To be specific, in each episode, $N$ classes are randomly sampled from $D_{test}$ firstly, and then $K$ instances for each of $N$ classes are sampled to constitute $D_{test-support}=\left\{ \left( x_{ij}^{s},y_{i}^{s} \right) ;i=1,...,N,\ j=1,...,K \right\}$. We usually call it $N-way$ $K-shot$ in FSL task. Besides, we randomly sample $R$ instances from the remaining samples of those $N$ classes to construct $D_{test-query}\left\{ \left( x_{k}^{q},y_{k}^{q} \right) ;k=1,...,R \right\}$. Likewise, we can also acquire $D_{train-support}$ and $D_{train-query}$ in a similar way, but $N\_for\_train$ is usually larger than $N\_for\_test$, which is a common practice in FSL. A data example of a 3-way 2-shot scenario of FSRC is given in Table \ref{tab:tab1}, where $N = 3$, $K = 2$ and $R = 1$.

As a result, FSRC can be defined as a task to predict the relation $y^q$ of a query instance $x^q$ in the query set $Q$, given a relation set $\mathcal{R}=\left\{ r_i;i=1,...,N \right\}$ and the support set $S$. As we consider $N = 5$ or 10, and $K = 5$ or 10 in this work, the number $N$ of classes and the number $K$ of instances are usually so small that the FS model has to be trained from the few instances in the support set and then use the trained model to predict the relation for any given query instance. In addition, each instance in the dataset is composed of $T$ words, which includes the mentioned entity pair $(h, t)$.

\begin{figure*}
\centering %图片居中
\includegraphics[height=12.0cm, width=0.95\textwidth]{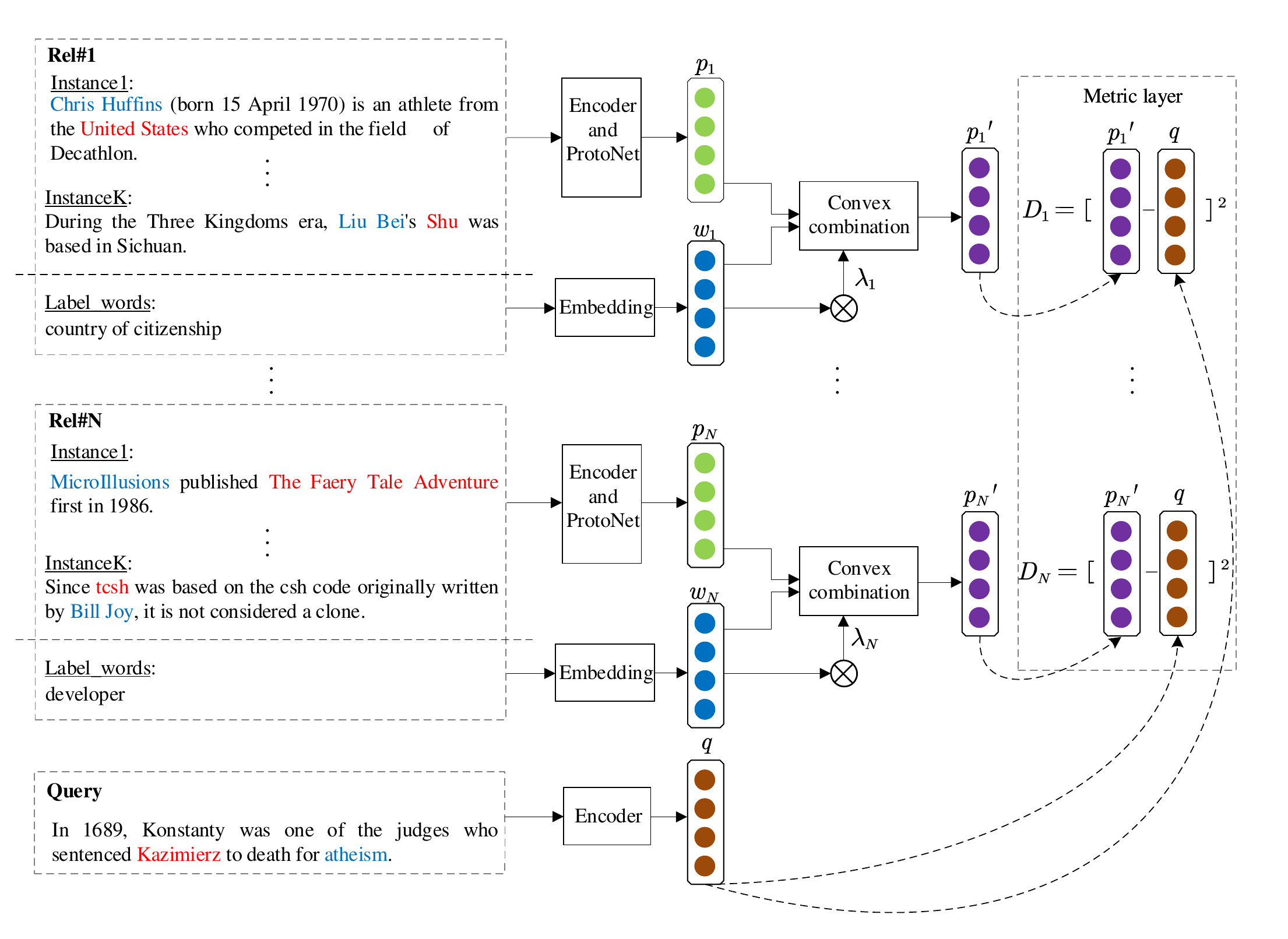} %插入图片，[]中设置图片大小，{}中是图片文件名
\caption{
The architecture of our proposed model for FSRC, where the input sample, instance encoder, prototypical networks and our proposed adaptive mixture mechanism are shown from left to right. The final category prototype is a convex combination of the semantic feature representation of samples in support set and label words, its coefficients are conditioned on the semantic label embedding. } %最终文档中希望显示的图片标题
\label{Fig_1}
\end{figure*}

\section{Method}
In this section, we present the details of the proposed adaptive prototypical networks with label words and joint representation learning (APN-LW-JRL) for FSRC. As shown in Fig. \ref{Fig_1}, all the input samples, both in the support set and in the query set, will be embedded to feed into the encoder to obtain their feature vector, while the label words are only expressed through the embedding layer. Then the class prototype can be attainable via inputting the represented samples in each support set into the prototypical networks. After that, we will incorporate the embedded label words into the class prototype by our proposed adaptive mixture mechanism to make improvements. Lastly, we will explain the specific joint representation learning method that we proposed, in an effort to perform effective end-to-end learning between input samples and output prediction results.

\subsection{Instance encoder}
The instance encoder mainly consists of two layers: embedding layer and instance encoding layer. The former aims to vectorize the input text, while the latter is meant to perform feature extraction on the vectorized text.

\subsubsection{Embedding Layer}
\
\newline
\indent Given an instance $x=\left\{ w_1,w_2,...,w_T \right\}$ with $T$ words mentioning two entity $(h, t)$, each word is represented by a real-valued vector $\mathbf{w}_i$ via the pretrained embedding matrix $V_w\in\mathbb{R}^{d_w\times \left| V \right|}$, where $\left| V \right|$ denotes the size of corpus vocabulary $V$ and $d_{w}$ means the dimension of word embedding. Furthermore, the positional features of each word in the sentence, i.e., the order of each word appears in the sentence, is very crucial for the extraction of text information. Especially in the task of RC, the closer the words are to the entities, the more informative the words are. For this reason, the position embeddings \cite{zeng2014relation} are introduced in our embedding layer, which we used to transform the relative distances between each word in the sentence and the two entities into vectors generated by looking up a randomly initialized position embedding matrix $V_p\in \mathbb{R}^{d_p\times \left| P \right|}$.

Finally, an ultimate input embedding for each word is achieved by concatenating word embedding and position embedding that can be denoted as $\mathbf{e}_{\boldsymbol{i}}=\left[\mathbf{w}_i;\mathbf{p}_{1i};\mathbf{p}_{2i} \right]$, where $\mathbf{e}_{\boldsymbol{i}}\in \mathbb{R}^d$ and $d=d_w+d_p\times 2$.

\subsubsection{Encoding Layer}
\
\newline
\indent In this work, CNN is adopted as our context encoder to obtain the hidden embedding of each word since it is very popular in most NLP tasks. After a convolution kernel with a window size $u$ acted on the input embedding $\left\{ \mathbf{e}_1,...,\mathbf{e}_T \right\}$, the hidden annotations will be as follows:
\begin{align}\label{2}
\mathbf{h}_i=CNN\left(\mathbf{e}_{i-\frac{u-1}{2}},...,\mathbf{e}_{i+\frac{u-1}{2}} \right)
\end{align}

Then to determine the most useful feature in each dimension of the feature vectors, we perform a max pooling operation on these hidden annotations.
\begin{align}\label{3}
\mathbf{x}=\max \left\{ \mathbf{h}_1,...,\mathbf{h}_T \right\}
\end{align}

Finally, we define the above two layers as a comprehensive function in Eq. (\ref{4}) for simplicity, where $\phi$ is the networks parameters to be learned.
\begin{align}\label{4}
\mathbf{x}=f_{\phi}\left( x \right)
\end{align}

\subsection{Prototypical networks}
The prototypical networks are simpler and more efficient than the meta-learning approaches in the FSL task, and they produce the state-of-the-art results even without sophisticated extensions developed for matching \cite{snell2017prototypical}. It is based on the idea that each class can be represented by means of its examples in a representation space learned by a neural network, which is expressed by Eq. (\ref{5}). $p_i$ is defined as class prototype for relation $r_i$ and $x_{ij}^{s}$ is the embedding of the $j$-th sentence of relation $r_i$ in the support set,  where $K$ sentences in total are included for each class.
\begin{align}\label{5}
p_i=\frac{1}{K}\sum_{j=1}^K{f_{\phi}\left( x_{ij}^{s} \right)}
\end{align}

After the prototypes for classes are calculated from supporting instances, we can classify a query instance by comparing the distance between the query instance and each prototype  based on a certain distance metric. Actually, given a distance function $D\left( \cdot \right)$, a distribution over classes is produced by prototypical networks for a query point $x^q$ based on a softmax function over distances to the prototypes in the embedding space. $\left| \mathcal{R} \right|$ denotes the total categories in the relation set.
\begin{align}\label{6}
p_{\theta}\left( y=r_i\left| x^q \right. \right) =\frac{\exp \left( -D\left( f_{\phi}\left( x^q \right) ,p_i \right) \right)}{\sum\limits_{l=1}^{\left| \mathcal{R} \right|}{\exp \left( -D\left( f_{\phi}\left( x^q \right) ,p_l \right) \right)}}
\end{align}

As indicated in \cite{snell2017prototypical}, there are multiple choices for the distance metric, among which the Euclidean distance is the most appropriate. Hence, we also adopt the Euclidean distance in this work.

\subsection{Adaptive prototypical networks}
In addition to the attractive simplicity, another reason why metric-based FSL approaches are so attractive is that they are able to learn a good embedding space, where samples from the same class are clustered together while samples from different classes are far away from each other. Afterwards, a new sample from the unseen category can be recognized directly through the distance metric within the learned embedding space. Upon that, it is very crucial to learn a discriminative embedding space in metric-based FSL tasks. In this section, we will interpret in detail how we perform adaptive technologies based on the prototypical networks by means of adaptive label information mixture mechanism and adaptive joint training for class prototype.

\begin{figure*}
\centering %图片居中
\includegraphics[height=10.0cm, width=0.8\textwidth]{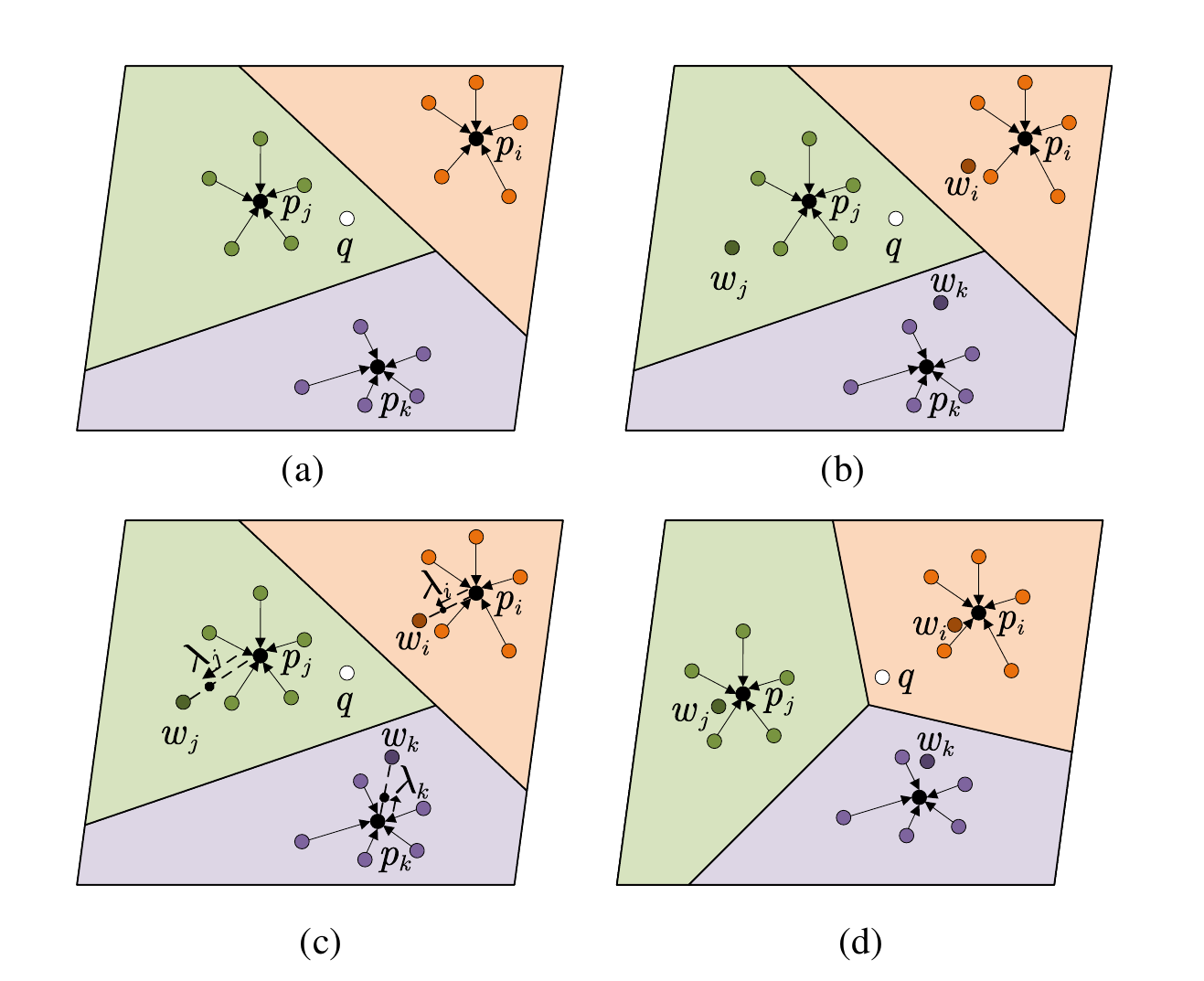} %插入图片，[]中设置图片大小，{}中是图片文件名
\caption{
The illustration of the proposed adaptive mixture mechanism. Suppose there is a query sample belonging to category $i$. (a) Originally, the prototype closest to the query sample $q$ is $p_j$. (b) Take the semantic feature of label words into account. (c) The location distribution of the class prototype has modified through the implementation of the mixture mechanism. (d) After the location distribution is updated, the closest prototype to the query is now the category $i$, correcting the classification.} %最终文档中希望显示的图片标题
\label{Fig_2}
\end{figure*}

\subsubsection{Adaptive Label Information Mixture Mechanism}
\
\newline
\indent As prototypical networks classify a query by calculating the distance between the mapping of each category, the sample distribution learned by the network model will not be really satisfactory when the two categories are particularly similar. Therefore, the classification performance may be affected with the more complicated textual features.

Thus, to make the spatial mapping of similar categories very distinctive, we hypothesize that semantic information of both samples and labels can be useful to make full use of the sample resources for FSL. Moreover, it is desirable to adaptively express the class prototypes, given different scenarios of categories. As shown in Fig. \ref{Fig_2}, we augment prototypical networks to model the new prototype representation as a convex combination of two pieces of information. For each category $i$, the new prototype is computed as follows.
\begin{align}\label{7}
p_i'=\lambda _i\cdot p_i+\left( 1-\lambda _i \right) \cdot c_i,
\end{align}
where $\lambda _i$ is the adaptive mixture coefficient that depends on the representation of category information to ensure that the calculated class prototype is compatible and not redundant.

\begin{align}\label{8}
\lambda _i=\frac{1}{1+\exp \left( -h\left( c_i \right) \right)}
\end{align}

\begin{align}\label{9}
c_i=g\left( w_i \right)
\end{align}

The label embedding for each class is first obtained by pretrained embedding matrix $V_w\in \mathbb{R}^{d_w\times \left| V \right|}$. Then we utilize an adaptive mixing network $h\left( \cdot \right)$, such as a fully connected network to infer the coefficient $\lambda _i$. This way, similar classes in the feature embedding space can be better separated. In addition, we use label description instead of label word information in the proposed adaptive mixture mechanism, which will be further discussed in Section \ref{sec:effects}.

\subsubsection{Adaptive joint learning for class prototype}
\label{sec:Adaptive joint training}
\
\newline
\indent The key issue of FSL is to learn to generalize and the performance of prototypical network for FSRC largely depends on the spatial distribution of sentence embeddings. To realize the full potential of metric based methods for FS, we propose an adaptive joint learning method based on the margin principle to learn a more discriminative metric space and improve the generalization capacity of model as well. This is achieved by combining the objective functions of representation learning and classification to get better feature representations for each instance while improving the classification performance. This way, the model is able to separate samples from different categories as farther as possible in the metric space.
\begin{align}\label{10}
\mathcal{L}_{representation}=\sum_{a=1}^{\left| S \right|+\left| Q \right|}{\left[ m+D_{\max}^{a,p}-D_{\min}^{a,n} \right] _+}
\end{align}

\begin{align}\label{11}
\mathcal{L}_{Cross-Entropy}=-\sum_{q=1}^{\left| Q \right|}{\log \left( p_{\theta}\left( y=r_i\left| x^q \right. \right) \right)}
\end{align}

\begin{align}\label{12}
\mathcal{L}_{Joint}=\mathcal{L}_{Cross-Entropy}+\alpha *\mathcal{L}_{representation}
\end{align}
Specifically, based on the attributes of the FSL tasks, we use the triple loss objective function introduced in Section \ref{sec:Triplet Loss in Face Recognition} as the objective function of representation learning, which consists of triplets of the anchor, the positive and the negative. Here, we use the Euclidean distance as our distance metric, denoted as $D$. For each anchor in the triplet loss, the difference to the vanilla method is that we select the farthest positive instance (denoted as $D_{\max}^{a,p}$) and the nearest negative instance (denoted as $D_{\min}^{a,n}$) to each sample, which are called hardest triplets. Particularly, we enriched the components of anchors that include all samples of the support set and query set in each episode. In Eq. (\ref{10}), $\left| S \right|+\left| Q \right|$ is the total triplets in each episode, and $m$ is a hyperparameter that controls the interval between categories. In Section \ref{sec:effects}, we further analyze the situation in which the triplet is composed of an anchor and its related class prototype. By combining the classification objective function and the representation learning objective function as shown in Eq. (\ref{12}), our model is able to take into account both classification error and representation quality at the same time during the training. In the loss function (\ref{12}), $\alpha$  is a hyperparameter controlling the weight of representation learning in the overall objective function. The overall framework of the proposed algorithm is summarized as Algorithm \ref{Algorithm 1}.

\begin{algorithm}[htbp]\footnotesize{
\caption{Adaptive prototypical networks with label words and joint learning for an episode in few-shot relation classification} \algblock{Begin}{End}
\label{Algorithm 1}
\renewcommand{\algorithmicrequire}{\textbf{Input:}}
\renewcommand{\algorithmicensure}{\textbf{Output:}}
\begin{algorithmic}[1]
 \Require
 Support set $S_i=\left\{ \left( \left. x_{ij}^{s},y_{i}^{s} \right) \right. \right\}$, query set $Q=\left\{ \left( x_{k}^{q},y_{k}^{q} \right) \right\}$, class-words $w_i$, $i=1,...N$, $j=1,...K$, $\mathcal{R}=\left\{ r_i;i=1,...,N \right\}$
 \Ensure
The joint learning objective function $\mathcal{L}_{Joint}$ in an episode.
\For {all samples $x_{ij}^{s}$ in class $i$}
\State Obtain the support prototype for class $i$ by feeding the support instances into the prototypical networks in Eq. (\ref{5});
\State  Obtain the semantic vector for class $i$ by feeding its words into a word embedding model in Eq. (\ref{9});
\State Compute the coefficient of the semantic vector in Eq. (\ref{8});
\State Obtain the class prototype for class $i$ in Eq. (\ref{7});
\State Calculate the probability that belongs to class $i$ according to the distance metric:
$p_{\theta}\left( y=r_i\left| x^q \right. \right) =\frac{\exp \left( -D\left( f_{\phi}\left( x^q \right) ,p_i' \right) \right)}{\sum\limits_{l=1}^{\left| R \right|}{\exp \left( -D\left( f_{\phi}\left( x^q \right) ,p_l' \right) \right)}}$
\State Compute the representation loss by finding the farthest positive instance and the nearest negative instance to each sample in the support set and query set in Eq. (\ref{10});
\State Obtain the final objective function by combining the cross-entropy of the classification objective function Eq. (\ref{12});
\EndFor
%\State Return $P_0$
\end{algorithmic}}
\end{algorithm}

\section{Experimental settings}
We describe the experimental settings in this section, beginning with introducing the dataset and learning configurations, followed by detailed implementation details of the models.

\subsection{Dataset and learning configurations}
In this work, we evaluate our model on the only FSRC dataset reported in the literature, FewRel, which is first generated by distant supervision and then filtered by crowdsourcing to remove noisy annotations. It consists of 70000 instances on 100 relations derived from Wikipedia, and each relation includes 700 instances. And both head and tail entities are marked in each sentence. Note, however, that only 80 relations are available, of which 48 are used for the training set, 12 for the verification, and 20 for test in the following comparative studies. In our experiments, we investigate four few-shot learning configurations, namely 5 way 1 shot, 5 way 5 shot, 10 way 1 shot, and 10 way 5 shot, which are the same for the proposed algorithm and the compared baselines. All results are averaged over ten independent runs.

\subsection{Implementation details}

\begin{table}[]
 \centering
 \caption{Hyperparameters of the models built in our experiments.}
\begin{tabular}{c|c|c}
\hline
\textbf{Component}            & \textbf{Parameter}                 & \textbf{Value} \\ \hline
\multirow{2}{*}{Embedding}    & Word embedding dimension $d_w$     & 50             \\ \cline{2-3}
                              & Position embedding dimension $d_p$ & 5              \\ \hline
\multirow{3}{*}{Encoder}      & Hidden layer dimension $d_h$       & 230            \\ \cline{2-3}
                              & Convolutional Window Size $u$      & 3              \\ \cline{2-3}
                              & Max length $T$                     & 40             \\ \hline
\multirow{2}{*}{Joint loss}   & Margin $m$                         & 0.5            \\ \cline{2-3}
                              & Alpha $\alpha$                             & 1              \\ \hline
\multirow{3}{*}{Optimization} & Initial Learning Rate              & 0.1            \\ \cline{2-3}
                              & Weight Decay                       & 10-5           \\ \cline{2-3}
                              & Dropout rate                       & 0.2            \\ \hline
\end{tabular}
\label{tab:tab2}
\end{table}

All hyperparameters of each part of our model are listed in Table \ref{tab:tab2}. We tune the rest hyperparameters of all models by grid search using the validation set, and all models are trained on the training set. Finally, the models achieving the best validation performance are saved to be tested on the test set. It has been found in \cite{munkhdalai2017meta} that feeding more classes to the models may achieve better performances than using the same configurations at both training and testing stages. Accordingly, we set $N = 20$ to construct the support set in each training episode. Actually, the training episode is built by first randomly selecting a subset of relations from the training set, then sampling a subset of instances within each selected relation to build the support set, and the remainder is sampled to build the query set. For fair comparisons, we use the same word embeddings pre-trained by Glove \cite{pennington2014glove} consisting of 6B tokens and 400K vocabularies, and the dimension of word embeddings is 50. In addition, the position embedding dimension of a word is add up to 10, the maximum length of each instance is 40, and the same encoder is used for both support and query instances. All experiments are based on the $N-way K-shot$ setting and we employ the mini-batch stochastic gradient descent (SGD) to solve the optimization problem.

\section{Results and analysis}
In this section, we will demonstrate the effectiveness of the proposed APN-LW-JRL on the FSRC task, followed by ablation studies. Finally, the visualization of sentence embeddings, the extension of support set and case study are respectively presented to show the validity of the proposed model.

\subsection{Compared algorithms}
To verify the effectiveness of APN-LW-JRL, we compare it with the following most state-of-the-art algorithms for FSRC:

Meta Network \cite{munkhdalai2017meta}: A novel meta-learning method that combines slow weights and fast weights for prediction.
\begin{itemize}
\item GNN \cite{garcia2017few}: A graph neural network that encodes each instance in support set and query set as a node in the graph for FSL.

\item SNAIL \cite{mishra2017simple}: A meta-learning model that uses temporal convolutional neural networks and attention mechanisms to quickly learn from past experience.

\item ProNet \cite{snell2017prototypical}: A model based on metric learning, which assumes that each class can be represented by a prototype.

\item Pro-HATT \cite{gao2019hybrid}: A hybrid attention-based prototypical network consists of instance-level and feature-level attention schemes.
\end{itemize}

\subsection{Comparative results}

\begin{table*}[]
 \centering
 \caption{Accuracies ($\%$) of different models on the divided FewRel test set under four different settings. Here, APN stands for adaptive prototypical networks, LW for label words, and JRL for joint representation learning.}
\begin{tabular}{c|cccc}
\hline
\textbf{Model}        & \textbf{5 Way 1 Shot} & \textbf{5 Way 5 Shot} & \textbf{10 Way 1 Shot} & \textbf{10 Way 5 Shot} \\ \hline
Meta Network          & $57.89\pm0.63$            & $71.02\pm0.45$            & $42.01\pm0.12$             & $53.87\pm0.32$                                 \\
GNN                   & $67.30\pm0.91$            & $79.47\pm0.61$            & $52.84\pm0.58$             & $69.63\pm0.21$                                 \\
SNAIL                 & $66.28\pm0.24$            & $78.07\pm0.12$            & $53.72\pm0.10$             & $62.88\pm0.33$                                 \\
ProNet                & $70.63\pm0.55$            & $85.10\pm0.22$            & $57.52\pm0.91$             & $74.74\pm0.11$                                  \\
Pro-HATT              & $71.76\pm0.63$            & $86.01\pm0.33$            & $59.22\pm0.21$             & $75.21\pm0.32$                                  \\ \hline
APN-LW                & $72.57\pm0.44$            & $86.29\pm0.15$            & $59.44\pm0.35$             & $76.32\pm0.05$                                  \\
APN-LW-JRL            & $\mathbf{73.45\pm0.83}$   & $\mathbf{87.27\pm0.55}$   & $\mathbf{61.02\pm0.61}$    & $\mathbf{77.69\pm0.22}$                     \\ \hline

\end{tabular}
\label{tab:tab3}
\end{table*}

The comparative results are presented Table \ref{tab:tab3}. From these results, we can make the following observations:
\begin{itemize}
\item APN-LW-JRL outperforms ProNet under the four different FSL settings, which is a simple but powerful baseline for few-shot learning and is also the basis of our model. Comparing ProNet and APN-LW, APN-LW and APN-LW-JRL, we can clearly see the effectiveness of introducing label words and joint training of classification and representation.

\item Strikingly, APN-LW-JRL also achieves better results compared with the other prototypical networks-based models, Pro-HATT, verifying the feasibility of considering prototypical networks in an adaptive way, even better than integrating various attention mechanisms. On the one hand, the adaptive label information mixture mechanism proposed in this work can leverage the information advantages of two aspects and adjust its focus accordingly. On the other hand, benefiting from adaptive joint training for class prototype, our model allows to perform end-to-end learning between the input instances and the desired embedding space.

\item APN-LW-JRL has shown significantly better results than Meta Network, GNN and SNAIL, which are the state-of-the-art learning models. This indicates that APN-LW-JRL is effective and the two adaptation mechanisms we introduced both make contributions to improve the performance.
\end{itemize}

\subsection{Effects of adaptive prototypical networks}
\label{sec:effects}
\begin{table*}[]
 \centering
 \caption{Accuracies ($\%$) of all models for ablation study on the divided FewRel test set under four different settings.}
\begin{tabular}{c|cccc}
\hline
\textbf{Model}      & \textbf{5 Way 1 Shot} & \textbf{5 Way 5 Shot} & \textbf{10 Way 1 Shot} & \textbf{10 Way 5 Shot} \\ \hline
APN-LD      & $71.78\pm0.31$        & $85.82\pm0.20$        & $59.20\pm0.11$         & $75.70\pm0.41$         \\
APN-LW      & $72.57\pm0.44$        & $86.29\pm0.15$        & $59.44\pm0.35$         & $76.32\pm0.05$         \\
APN-LD-JRL  & $73.18\pm0.22$        & $86.97\pm0.31$        & $60.67\pm0.52$         & $77.09\pm0.16$         \\ \hline
ProNet-PJRL & $71.63\pm0.35$        & $86.20\pm0.72$        & $58.92\pm0.15$         & $76.23\pm0.66$         \\
APN-LW-PJRL & $72.56\pm0.11$        & $86.55\pm0.56$        & $59.80\pm0.22$         & $76.99\pm0.61$         \\
ProNet-JRL  & $72.98\pm0.23$        & $86.62\pm0.45$        & $60.25\pm0.12$         & $76.43\pm0.35$         \\ \hline
APN-LW-JRL  & $\mathbf{73.45\pm0.83}$   & $\mathbf{87.27\pm0.55}$  & $\mathbf{61.02\pm0.61}$    & $\mathbf{77.69\pm0.22}$     \\ \hline
\end{tabular}
\label{tab:tab4}
\end{table*}

\begin{figure}
\centering %图片居中
\includegraphics[height=8.0cm, width=0.5\textwidth]{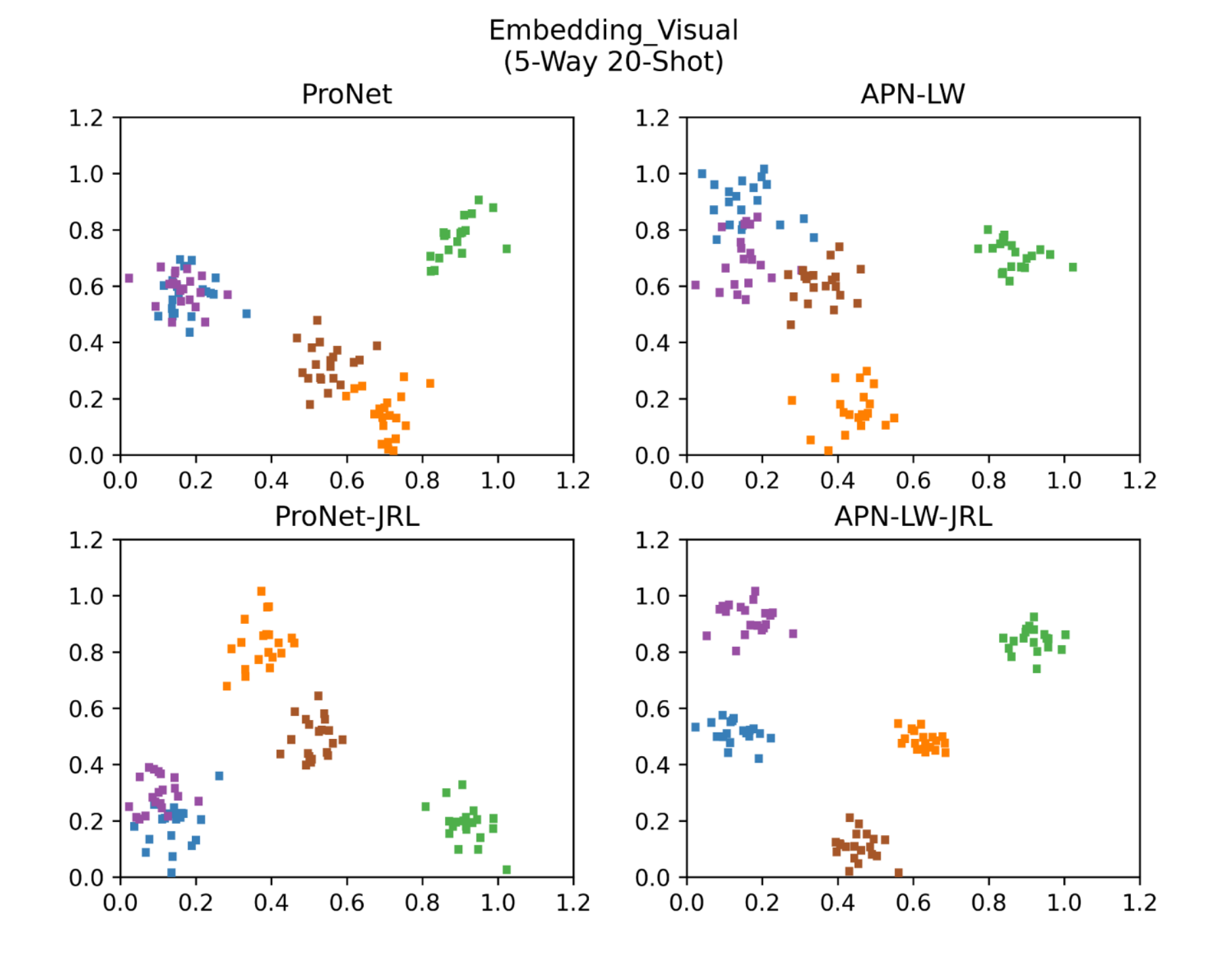} %插入图片，[]中设置图片大小，{}中是图片文件名
\caption{
 Visualization of sentence embeddings of a 5-way-20-shot scenario in the divided FewRel test set, where one point represents one instance and different colors correspond to different categories. } %最终文档中希望显示的图片标题
\label{Fig_3}
\end{figure}

In this section, ablation studies were conducted to evaluate the contributions of the individual model components. Variants of the proposed algorithm, APN and ProNet are designed to verify the respective function of the adaptive label information mixture mechanism and adaptive joint representation learning for class prototype.
\begin{itemize}
\item APN-LD: The proposed adaptive label information mixture mechanism based on label description.

\item APN-LD-JRL: The proposed adaptive label information mixture mechanism based on label description with adaptive joint representation learning based on the sample triplets.

\item ProNet-JRL: ProNet \cite{snell2017prototypical} model with the proposed adaptive joint representation learning based on the sample triplets.

\item ProNet-PJRL: ProNet \cite{snell2017prototypical} model with the proposed adaptive joint representation learning based on the prototype triplets.

\item APN-LW: The proposed adaptive label information mixture mechanism based on label words.

\item APN-LW-PJRL: The proposed adaptive label information mixture mechanism based on label words with adaptive joint representation learning based on the prototype triplets.

\item APN-LW-JRL: The proposed adaptive label information mixture mechanism based on label words with adaptive joint representation learning based on the sample triplets.
\end{itemize}

Table \ref{tab:tab4} shows the performance of our model and its ablations on the divided FewRel test set. By comparing the three models of ProNet (in Table \ref{tab:tab3}), APN-LD and APN-LW, we can find that the adaptive label information mixture mechanism exhibits a positive influence on the expression of class prototypes. Here, LD denotes label description and its representation is obtained by the same encoder as the instance. Comparing APN-LD and APN-LW, we find that LD, which is much longer than LW, is not as good as LW. This is contrary to the intuitive belief that the longer the text, the richer information it contains. We take a closer look at the dataset and find that this might be attributed to the fact that the descriptions of some categories are too metaphysical. It may cause deviation or redundancy when the description is used to represent the label information. For example, the LW “residence” is more idiographic than the LD “the place where the person is or has been”, which helps to get a better class prototype. At the same time, the result of comparing APN-LD-JRL and APN-LW-JRL can also be used as convincing evidence in support of this argument.

In addition, we also compare ProNet and ProNet-JRL, APN-LW and APN-LW-JRL, respectively, from which we can conclude that the adaptive joint representation learning method we proposed demonstrates an excellent end-to-end deep metric learning capability. PJRL here means that for each sample as an anchor, the farthest class prototype with the same label is chosen as positive, while the nearest class prototype with a different label is chosen as negative to form the triplets. Clearly, the generalization performance of selecting the triples with the class prototype as the target is not as strong as the sample-targeting, because the selection of samples to form the triples is more extensive, which can be concluded by comparing with ProNet-PJRL and ProNet-JRL, APN-LW-PJRL and APN-LW-JRL.

Furthermore, we visualize an episode set of sentences from the divided FewRel test set to gain more insight into the model performance. It can be seen from Fig. \ref{Fig_3} that as adaptive mixture mechanism and adaptive joint representation learning play an increasing role, APN-LW-JRL is able to learn more discriminative sentence embeddings, making it more capable of distinguishing confusing data than those of the previous ablation models. We attribute this capability to the rich feature expression of the adaptive mixture mechanism and the moderate spatial embedding learning of adaptive joint representation learning, which increases the distance between different classes and reduces the distance within the same class.

\subsection{Extension of support set}

\begin{figure*}
\centering %图片居中
\includegraphics[height=6.5cm, width=0.9\textwidth]{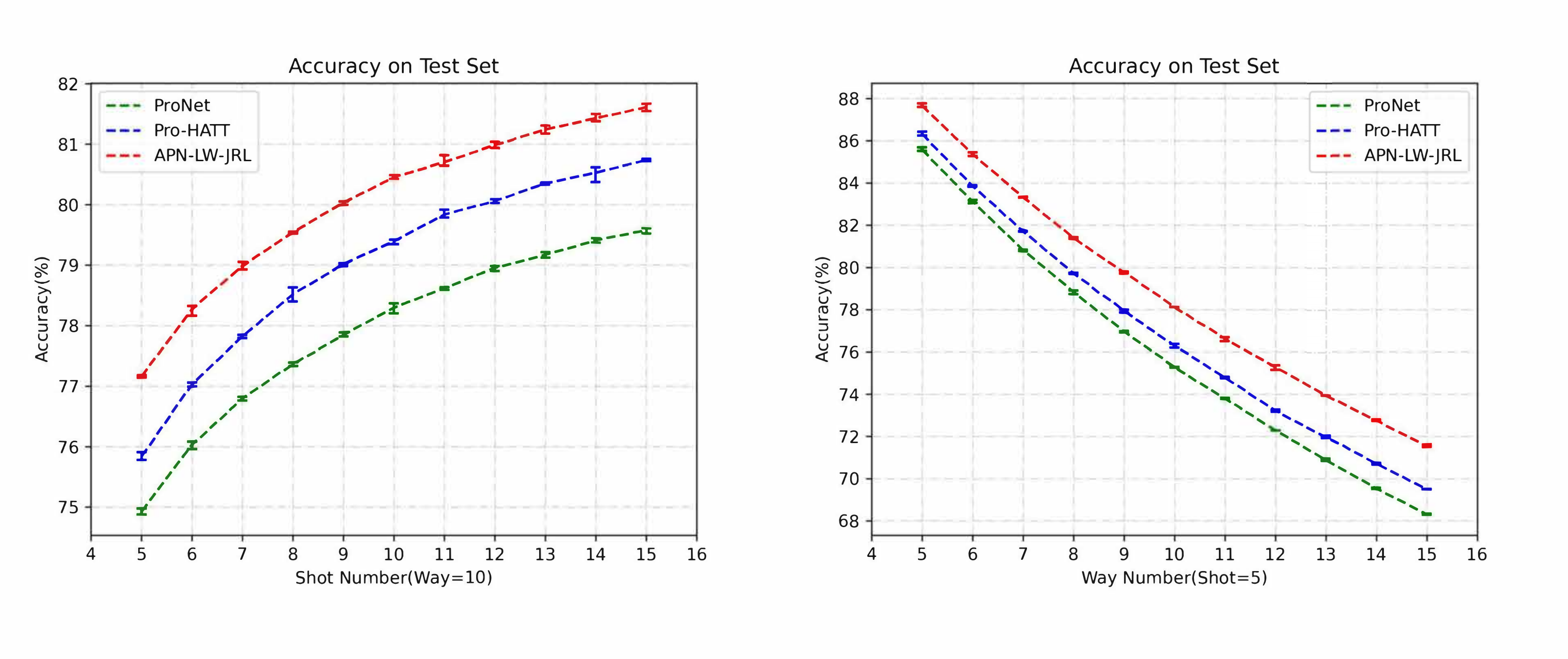} %插入图片，[]中设置图片大小，{}中是图片文件名
\caption{
 Comparison of performance of Proto, Pro-HATT and APN-LW-JRL for different data extensions of the support set on the divided FewRel test set. On the left, the value of the way is fixed to 10, and the accuracy changes with the shot. On the right, the value of the shot is fixed to 5, and the accuracy changes with the way. Each small vertical line segment on the curve represents an error interval of multiple experiments. } %最终文档中希望显示的图片标题
\label{Fig_4}
\end{figure*}

\begin{table*}[]
 \centering
 \caption{Accuracies ($\%$) of Proto, PHATT and HAPN in different data extensions of the support set on the divided FewRel test set.}
\begin{tabular}{c|ccccccccccc}
\hline
                   & \multicolumn{11}{c}{\textbf{10-way K-Shot Accuracy ($\%$)}}                           \\ \hline
K         & 5     & 6     & 7     & 8     & 9     & 10    & 11    & 12    & 13    & 14    & 15    \\ \hline
ProNet    & 74.92 & 76.02 & 76.79 & 77.36 & 77.84 & 78.29 & 78.61 & 78.95 & 79.16 & 79.41 & 79.57 \\
Pro-HATT  & 75.83 & 77.02 & 77.81 & 78.51 & 79.01 & 79.39 & 79.83 & 80.06 & 80.34 & 80.52 & 80.73 \\
APN-LW-JRL & \textbf{77.16}& \textbf{78.27}& \textbf{78.98}& \textbf{79.54}& \textbf{80.03}& \textbf{80.46}& \textbf{80.70}& \textbf{80.98}& \textbf{81.24}&\textbf{81.43} &\textbf{81.61} \\ \hline
                   & \multicolumn{11}{c}{\textbf{N-way 5-Shot Accuracy ($\%$)}}                           \\ \hline
N         & 5     & 6     & 7     & 8     & 9     & 10    & 11    & 12    & 13    & 14    & 15    \\ \hline
ProNet    & 85.58 & 83.12 & 80.82 & 78.84 & 76.96 & 75.27 & 73.78 & 72.28 & 70.87 & 69.53 & 68.31 \\
Pro-HATT  & 86.31 & 83.86 & 81.71 & 79.72 & 77.94 & 76.28 & 74.79 & 73.22 & 71.95 & 70.71 & 69.50 \\
APN-LW-JRL & \textbf{87.66}& \textbf{85.37}& \textbf{83.33}& \textbf{81.40}& \textbf{79.77}& \textbf{78.13}& \textbf{76.62}& \textbf{75.29}& \textbf{73.95}& \textbf{72.75}& \textbf{71.54}\\ \hline
\end{tabular}
\label{tab:tab5}
\end{table*}

It is recognized that more support instances can provide more useful information to the prototype vector, but more noise may be added in as well. In this section, to study how the number of categories and samples in the support set will affect the performance of different models, we define an extension of the support set as the additive value of accuracy. First, $K$ is changed from 5 to 15, when $N$ to 10.  Then, the range of $N$ is also 5-15 when $K$ is fixed to 5. As shown in Fig. \ref{Fig_4} and Table \ref{tab:tab5}, given the same extension setting, the proposed model shows much better performance than other benchmarks. It also verifies that the way we increase the data utilization is beneficial to improve the robustness and generalization capability of our model. %On this basis, we can accumulate as much annotated data as possible, thereby improving the performance of the task of FSRC in real life.

\subsection{Case studies}

\begin{table*}[]
 \centering
 \caption{Some specific results of relation classification examples from FewRel dataset. The head entity and the tail entity are marked as blue and red respectively in each instance.}
\begin{tabular}{llccc}
\hline
\textbf{Label}                                                                 & \textbf{Sentence}                                                                                                                                                                 & \textbf{ProNet}                                                         & \textbf{APN-LW}                                                           & \textbf{APN-LW-JRL}                                                       \\ \hline
\begin{tabular}[c]{@{}l@{}}"P1001":\\ Applies to jurisdiction\end{tabular}     & \begin{tabular}[c]{@{}l@{}}Macelod was a territorial electoral district for \\ the \textcolor[rgb]{1.00,0.00,0.00}{Legislative Assembly} of \textcolor[rgb]{0.00,0.50,1.00}{Northwest Territories},\\ Canada.\end{tabular}                          & \begin{tabular}[c]{@{}c@{}}Location of formation\\ (False)\end{tabular} & \begin{tabular}[c]{@{}c@{}}Applies to jurisdiction\\ (True)\end{tabular}  & \begin{tabular}[c]{@{}c@{}}Applies to jurisdiction\\ (True)\end{tabular} \\ \hline
\begin{tabular}[c]{@{}l@{}}"P463":   \\ Member of\end{tabular}                 & \begin{tabular}[c]{@{}l@{}}South Africa is part of the \textcolor[rgb]{0.00,0.50,1.00}{IBSA Dialogue Forum}, \\ alongside \textcolor[rgb]{1.00,0.00,0.00}{Brazil} and India.\end{tabular}                                                           & \begin{tabular}[c]{@{}c@{}}Part of   \\ (False)\end{tabular}            & \begin{tabular}[c]{@{}c@{}}Member of   \\ (True)\end{tabular}             & \begin{tabular}[c]{@{}c@{}}Member of   \\ (True)\end{tabular}            \\ \hline
\begin{tabular}[c]{@{}l@{}}"P937": \\ Work location\end{tabular}               & \begin{tabular}[c]{@{}l@{}}She lives in \textcolor[rgb]{0.00,0.50,1.00}{Montreal} and is the common - law partner \\ of novelist \textcolor[rgb]{1.00,0.00,0.00}{Rawi Hage}.\end{tabular}                                                           & \begin{tabular}[c]{@{}c@{}}Residence\\ (False)\end{tabular}             & \begin{tabular}[c]{@{}c@{}}Work location\\ (True)\end{tabular}            & \begin{tabular}[c]{@{}c@{}}Work location   \\ (True)\end{tabular}        \\ \hline
\begin{tabular}[c]{@{}l@{}}"P355": \\ Subsidiary\end{tabular}                  & \begin{tabular}[c]{@{}l@{}}Hotel Indigo competes with \textcolor[rgb]{1.00,0.00,0.00}{Starwood} 's W Hotels as \\ well as Andaz Hotels, \textcolor[rgb]{0.00,0.50,1.00}{Aloft Hotels} and Le Meridien \\ Hotels.\end{tabular}                       & \begin{tabular}[c]{@{}c@{}}Has part   \\ (False)\end{tabular}           & \begin{tabular}[c]{@{}c@{}}Subsidiary   \\ (True)\end{tabular}            & \begin{tabular}[c]{@{}c@{}}Subsidiary   \\ (True)\end{tabular}           \\ \hline
\begin{tabular}[c]{@{}l@{}}"P1435": \\    \\ Heritage designation\end{tabular} & \begin{tabular}[c]{@{}l@{}}Hampton Hill, John Thompson House, Twin Trees Farm, \\ and \textcolor[rgb]{1.00,0.00,0.00}{Willow Mill Complex} are listed on the \textcolor[rgb]{0.00,0.50,1.00}{National} \\ \textcolor[rgb]{0.00,0.50,1.00}{Register of Historic Places}.\end{tabular} & \begin{tabular}[c]{@{}c@{}}Instance of\\ (False)\end{tabular}           & \begin{tabular}[c]{@{}c@{}}Applies to jurisdiction\\ (False)\end{tabular} & \begin{tabular}[c]{@{}c@{}}Heritage designation\\ (True)\end{tabular}    \\ \hline
\end{tabular}
\label{tab:tab6}
\end{table*}

In order to better understand why the proposed model outperforms the compared ones, we selected a few examples from the divided FewRel test set and analyze their classification results. Table \ref{tab:tab6} shows five instances on which ProNet has failed while our model has correctly predicted. Take the second query instance in Table \ref{tab:tab6} as an example, ProNet predicts this incorrectly into “Part of”, while  our model classifies accurately. This instance is quite challenging since the expressions of these two relations are very similar. From these instances, we can conclude that it is of great importance to update both the way of class prototype generation and the learning target of the original prototypical networks, as done in our model.

\section{Conclusion and Future Work}
Metric learning based FSL task aims to learn a set of projection functions that take support and query samples from the target problem, which is heavily dependent on the quality of the embedding space. Therefore, to maximize the efficiency of existing information resources and achieve a better class prototype, we improve the prototypical networks in an adaptive manner, which consists of a label information mixture mechanism and a joint representation learning method for FSRC task. It can appropriately weigh the importance of support samples and label information to make adaptive adjustments and learn a more discriminative prototype representation. Comparative studies have been conducted on FewRel under different FS settings, from which we can conclude that our proposed APN-LW-JRL is successful in improving the classification performance of few-shot RC models.

In this work, label information of the data is made use of to improve the performance of the proposed APN-LW-JRL and the proposed algorithm is tested on the FewRel data set only. In the future, we will extend the model to zero-shot learning and validate its performance on additional text datasets. In addition, we are going to investigate the use of few shot learning for joint relation extraction of named entity recognition (NER) and relation classification (RC), and its application to medical and other real-world problems.

\bibliographystyle{IEEEtran}

%?Loading?bibliography?database
\small
\bibliographystyle{ieee}
%\bibliography{CASSreference}

\bibliography{cas-refs}

\end{document}